\documentclass[letterpaper]{article}
\usepackage{proceed2e}
\usepackage[margin=1in]{geometry}

\usepackage{times}
\usepackage{graphicx}
\usepackage{subfigure}
\usepackage{amsfonts} 

\usepackage{amsmath}  
\usepackage{nicefrac} 
\usepackage{url}
\usepackage{multirow} 
\usepackage[authoryear,round]{natbib}
\usepackage{fancyhdr}

\usepackage{xcolor,colortbl}

\newcommand{\mc}[2]{\multicolumn{#1}{|c|}{#2}}
\definecolor{Gray}{gray}{0.85}
\newcolumntype{a}{>{\columncolor{Gray}}c}
\newcolumntype{b}{>{\columncolor{white}}c}

\renewcommand{\headrulewidth}{0pt}

\DeclareMathOperator{\sign}{sign}

\title{Adversarial Training Versus Weight Decay}

\author{ {\bf Angus Galloway}\\
School of Engineering\\
University of Guelph, Canada\\
Vector Institute, Canada\\
\texttt{gallowaa@uoguelph.ca}
\And 
{\bf Thomas Tanay}\\
CoMPLEX\\
Dept.~of Computer Science\\
University College London, UK\\
\texttt{thomas.tanay.13@ucl.ac.uk}\\
\And 
{\bf Graham W.~Taylor}\\
University of Guelph, Canada\\
Canadian Institute for Advanced Research\\
Vector Institute, Canada\\
\texttt{gwtaylor@uoguelph.ca}\\
}

\begin{document}
\maketitle
\begin{abstract}
Performance-critical machine learning models should be robust to
input perturbations not seen during training. Adversarial training
is a method for improving a model's robustness to~\emph{some} perturbations by
including them in the training process, but this tends to exacerbate other
vulnerabilities of the model. The adversarial training
framework has the effect of~\emph{translating} the data with respect
to the cost function, while weight decay has a~\emph{scaling}
effect. Although weight decay could be considered a crude regularization
technique, it appears superior to adversarial training as it remains
stable over a broader range of regimes and reduces all generalization errors.
Equipped with these abstractions, we provide key baseline results and
methodology for characterizing robustness. The two approaches
can be combined to yield~\emph{one} small model that demonstrates
good robustness to several white-box attacks associated with different
metrics.\footnote{Source code and pre-trained models available
\url{https://github.com/uoguelph-mlrg/adversarial_training_vs_weight_decay}}
\end{abstract}

\section{Introduction}
Machine learning models have been shown to work well in a variety of
domains, for the equivalent of~\emph{laboratory conditions} in the
physical sciences. As engineers increasingly opt to deploy these models
in performance-critical applications such as autonomous driving or
in medical devices, it is necessary to characterize
how they behave in the field. In the aforementioned
applications, the confidence with which a prediction was made
can be even more important than the prediction itself.
Unfortunately, deep neural networks (DNNs) leave much to be
desired, typically offering little interpretability
to their predictions, unreliable confidence estimates,
and tending to fail in unexpected ways~\citep{biggio2013evasion, szegedy2014intriguing, goodfellow2015explaining, nguyen2015deep}.
The~\emph{adversarial examples} phenomenon afflicting machine learning
models has shed light on the need for higher levels of abstraction to
confer reliable performance in a variety of environments, where models may be
pushed to the limit~\emph{intentionally} by adversaries, or~\emph{unintentionally}
with out-of-distribution data. Although such
settings could be seen as violating the i.i.d assumption implicit in typical
machine learning experiments, raising this objection provides little comfort
when systems fail unexpectedly.

In this work, we compare the approach of augmenting training data with
worst-case examples~\citep{szegedy2014intriguing}, which can result in
somewhat mysterious effects~\citep{kurakin2017adversarial, tramer2018ensemble},
to weight decay, which is more mathematically
understood~\citep{neyshabur2015normbased, goodfellow2016deep}.\
There remains a widely held belief that adversarial
examples are linked to the linear behaviour of machine learning
models~\citep{goodfellow2015explaining}. We begin by developing stronger
intuition for linear models, finding that adversarial training
can be seen as translating the data w.r.t.~the loss, while weight decay
scales the loss. With these abstractions, we report a 40\% improvement in
absolute accuracy over a previous baseline, while incurring a modest 2.4\%
loss for the natural data.

We leverage the translation and scaling perspective to tune a low-capacity
model to be more robust for~\emph{medium to large} perturbations
bounded in the Hamming distance (or $L_0$ counting ``norm'') for the CIFAR-10
dataset~\citep{krizhevsky2009tiny} than another compelling defense against
white-box attacks~\citep{madry2018towards}. The same model
is competitive for $L_\infty$ bounded attacks, and outputs high-entropy
distributions over its softmax-probabilities for out-of-distribution
examples, such as~\emph{fooling images}~\citep{nguyen2015deep}.
The model seems more interpretable, and has two orders of magnitude fewer
parameters than the current baseline.
\section{Related Work}
\label{sec:related-work}
After two investigations of defenses against adversarial
examples~\citep{athalye2018obfuscated, uesato2018adversarial}, the only approach
that maintained a significant robustness improvement was based on adversarial
training. Many previous works attempt to
slow down attackers by manipulating or masking gradients at
test time, but these do nothing to improve generalization.
We refer the interested reader
to~\citet{carlini2017towards} and \citet{athalye2018obfuscated} for
concise summaries of common mistakes.~The iterative
projected gradient descent (PGD) method with a random start,
proposed by~\citeauthor{madry2018towards},~currently holds
a benchmark for robustness to perturbations w.r.t.~the
CIFAR-10 test set, for a
threat model in which an adversary may perturb any number of pixels by
$\pm8$ from their original integer values $\in$ [0, 255].
However, there is growing concern that this $L_\infty$
norm threat model is unrealistically
weak~\citep{sharma2017attacking,tramer2018ensemble,athalye2018obfuscated,galloway2018predicting}.

By slightly extending the threat model of~\citeauthor{madry2018towards},
the PGD defense has been found to perform
worse than an equivalent~\emph{undefended} model for MNIST~\citep{mnist98},
which is considered one of the easier datasets to defend on due to its low
dimensionality and high separability.~Weaknesses of the PGD defense for MNIST
include adding a constant to all pixels~\citep{galloway2018predicting}, rotating
images by $20^\circ$ prior to attack, or randomly flipping just~\emph{five}
pixels~\citep{tramer2018ensemble}.
An elastic-net black-box transfer attack~\citep{sharma2017attacking}
achieved higher success rates than the attacks considered by~\citeauthor{madry2018towards},~but with less distortion in
the same $L_\infty$ norm. For CIFAR-10, the
official~\citeauthor{madry2018towards}~defense fails
the~\emph{fooling images} test~\citep{nguyen2015deep} by classifying
with maximum confidence unrecognizable images initialized from
uniform noise~\citep{galloway2018predicting}.

A formal certificate of robustness based on adversarial training was
recently provided by~\cite{sinha2018certifiable}, and while their principled
approach is laudable, the threat model considered is weaker than that
of~\cite{madry2018towards} and exclusively deals with~\emph{imperceptibly}
small perturbations, which were not well defined.
The translation view of adversarial training proposed in this work offers
insight into the difficulty of using the framework with large perturbations,
and suggests that it may indeed be inherently limited to the small perturbation
regime.

In support of the threat model chosen by~\citeauthor{madry2018towards},
there are still few standards for assessing robustness,
and the $L_\infty$ metric facilitates convenient high-level comparisons based on
a single data-point.
To the best of our knowledge, the original use of $L_\infty$ for bounding
adversarial examples arose from a desire to mitigate perturbations
akin to steganography\footnote{The encoding of secrets in plain
sight, e.g.,~in an image.}.~Under $L_\infty$, the cumulative effect of changes to
each input dimension (i.e.~pixel) smaller than the least resolvable feature
in the digital representation of the training data (e.g.~with 8-bits), is to
be ``decoded'' such that any desired class is predicted with
high-confidence~\citep{goodfellow2015explaining}.

There are many types of perturbations models could be subject to,
therefore we intentionally depart from the traditional
security literature to explore trade-offs implicit in choosing a
specific threat model. We sweep perturbation magnitude through
the~\emph{full} dynamic range, or until accuracy is reduced to
less than chance, for attacks that are optimized for different $L_p$
metrics\footnote{For $p=0$, we mean
Hamming distance.}~(e.g.~$p \in {0,1,2,\infty}$).
For security, performance is no better than the worst of any particular
case, but depending on the application, practitioners may wish to choose
the model that maximizes area under all curves. Attacking until models
reach~\emph{zero} accuracy was proposed by~\citet{athalye2018obfuscated}
to rule out~\emph{gradient masking} as contributing to a false positive
defense. We find that additional insight can be gained even
between~\emph{chance} and~\emph{zero}, therefore we usually opt for the latter.

It was recently shown that regularizing~\emph{shallow} convolutional neural
networks with strong $L_2$ weight decay does mitigate the fooling
images attack, also known as ``rubbish'' examples, for the MNIST and CIFAR-10
datasets~\citep{galloway2018predicting}. For $L_\infty$ bounded perturbations
of natural CIFAR-10 data, these models perform more closely to the PGD-trained
WideResNet benchmark~\citep{madry2018towards} than the other defenses
tested by~\citet{athalye2018obfuscated}, despite having only 1\% as many
parameters.

Previous works briefly tested or acknowledged the role of
weight decay as a possible defense against adversarial
perturbations, but it is generally discarded along with
other ``regularization techniques'' as adversarial examples
are not typically viewed as an overfitting
problem~\citep{szegedy2014intriguing, goodfellow2015explaining}.
Follow on works generally refer to this preliminary
conclusion, and do not consider weight decay given a constraint
of not ``seriously harming'' classification accuracy on
benign inputs~\citep{papernot2016distillation}~\footnote{In a personal
communication with the first author, they clarified that adversarial examples
are partly due to overfitting, but that this is not the only explanation
for their existence, therefore traditional regularization methods are not
sufficient to mitigate adversarial examples. They did not intend to imply that
robustness should come at no cost of accuracy for the natural data.}.

We maintain that most adversarial examples observed in
practice are of the ``Type 2''---overfitting---variety in the taxonomy
of~\cite{tanay2016boundary}, and that ``serious harm'' is too
arbitrary for a~\emph{constraint}, and should probably be a criteria.
Other works used a small weight decay penalty concurrently with other
regularization techniques (e.g.~dropout) that exhibit different
behaviour, and where the contribution of the norm penalty is insignificant,
or where it is only applied to the last layer of a deep
network~\citep{cisse2017parseval}.

\citet{rozsa2016towards}~argue that correctly classified examples have a
small contribution in the cross-entropy loss, leading to ``evolutionary
stalling'' that prevents the decision boundary from flattening near the
training data. However, if we consider the high-dimensional geometry of
finite volume sub-manifolds corresponding to each class in a fixed dataset,
flatter shapes actually~\emph{minimize} the smallest adversarial distance,
while a manifold with constant positive curvature (e.g.~a
sphere)~\emph{maximizes} distance in all
directions.~\citeauthor{rozsa2016towards}~found that batch
adjusting network gradients (BANG) increases overfitting on
natural data, which they also address with dropout, further
complicating the experiments.
Both~\citeauthor{rozsa2016towards}
and~\citeauthor{cisse2017parseval}~evaluate on weak~\emph{single-step}
attacks that linearly approximate the loss of their deep models.
Success against linear attacks can be improved by making models
less linear, but non-linear models do not necessarily
generalize better. Non-linearity is defeated with iterative attacks which
can still yield examples that transfer, thus enabling black-box
attacks~\citep{tramer2018ensemble}.

\citeauthor{tramer2018ensemble}~found that adversarial training
on ImageNet often converges to a~\emph{degenerate global minimum},
leading to a kind of learned~\emph{gradient masking} effect.
Rather than becoming robust, it's conceivable that models may intentionally
contort their decision boundaries s.t.~\emph{weaker} adversarial
examples are crafted that are easy to
classify.~\citeauthor{tramer2018ensemble}~attempt to
prevent this by transferring examples from an ensemble of pre-trained
models, hence~\emph{decoupling} adversarial example generation from the
model being learned. They claim improved robustness to some black-box
transfer attacks, but at the expense of white-box performance.
The ensemble method was omitted from the analyses
of~\cite{athalye2018obfuscated}, as white-box robustness is strictly more
difficult than in the black-box setting. The technique was shown to not
hold in a more general black-box threat model using a query-efficient gradient
estimator~\citep{ilyas2017query, ilyas2018ensattack}, and a gradient-free
genetic algorithm~\citep{alzantot2018genattack}.

Our translation perspective can explain why decoupling adversarial
example generation is insufficient: degenerate global
minima are an inherent result arising from training on worst-case perturbations
with fixed magnitude, which fails to account for changing distances between
individual training data and the decision boundary as a model
learns. To perform adversarial training in a way that is mathematically
correct requires taking into account the distance of each example
from the class mean.

Recently, there has been a renewed interest in developing
knowledge in norm-based capacity control of linear and depth-limited
networks~\citep{bengio2006convex, neyshabur2015normbased} as an explanation for
generalization in deep learning~\citep{neyshabur2017exploring, tsuzuku2018lipschitz}.
On the other hand, there remains an operating
hypothesis in many recent works that the existence of adversarial examples
is linked to the linear behaviour
of machine learning models~\citep{goodfellow2015explaining}.~Norm-based
capacity control is well-characterized for linear models, letting us
explicitly deal with adversarial examples as a subset of
the generalization problem in this setting. Linear models are
thought to be highly interpretable, or at least more so than their deep
non-linear counterparts, therefore it can seem unintuitive that they may
experience adversarial examples. However, we caution against using this
premise to refute attempts at making models more interpretable~\emph{in general},
as linear models do not always suffer from adversarial examples.

Previous work has identified limitations with the linear interpretation for the
existence of adversarial examples, through manipulating hidden representations
in DNNs with feature adversaries~\citep{sabour2015adversarial}, the boundary
tilting perspective~\citep{tanay2016boundary}, and an insightful commentary on
high-dimensional geometry~\citep{dube2018high}.~Most similar to this work is
that of~\citeauthor{tanay2016boundary}, however they did not empirically
validate their perspective with DNNs, nor discuss adversarial training.
To limit further confusion around linearity, we begin by characterizing linear
models for binary classification, then evaluate DNNs on CIFAR-10.

\section{Robust Logistic Regression}
\label{sec:robust-logistic}

To support our rejection of the ``linearity hypothesis'',
we re-visit Section 5 from~\cite{goodfellow2015explaining} in which it
is motivated by observing that a linear classifier, logistic regression, is
vulnerable to adversarial examples for MNIST ``3'' versus ``7'' binary
classification. They introduce a single-step attack known as the~\emph{fast
gradient method} (FGM), of which the~\emph{sign} variant (FGSM) is the strongest
possible attack against a linear model when an adversary is constrained in the
$L_\infty$ norm.
Perturbations are scaled by a constant $\epsilon$ when using the
$L_\infty$ norm, where $\epsilon$ can always be interpreted as a fraction of
the input range which has been normalized between [0, 1] when it appears in
decimal form. In this setting, \citeauthor{goodfellow2015explaining}\ reported a
$99\%$ test error rate for an FGSM perturbation scaled by $\epsilon=0.25$,
which we denote ``the baseline'' in Sections~\ref{sec:robust-logistic}
and~\ref{sec:adv-vs-wd-linear-models}. To the best of our knowledge, no
updated result has since been reported for this
experiment.~\citeauthor{goodfellow2015explaining}~later compared the
regularizing effects of adversarial training on FGSM examples to $L_1$ weight
decay, which we re-visit in Section~\ref{sec:adv-vs-wd-linear-models}.

For the experiments in this section, we use a parameter initialization
scheme that consists of subtracting the average MNIST ``7'' from the
average ``3'', which we refer to as ``expert initialization''.
By simply using this initialization, we obtain a $39\%$ absolute percentage
improvement over the baseline, but the model underfits
with $\approx 5.2\%$ error on the clean test set. These are approximately the
resulting set of weights after empirical risk
minimization (ERM) fully converges when training from scratch with $L_2$ weight
decay, if one accounts for minor differences due to stochastic optimization
with a particular batch size, learning rate, etc. In addition to conferring
good robustness, we use this expert scheme to facilitate reproducibility.
We depict weights, and perturbations crafted by FGSM in
Figure~\ref{fig:mnist_three_seven}.~\citeauthor{goodfellow2015explaining}~comment
on Figure~\ref{fig:logreg_weights_sign} in their original caption
that the perturbation is not readily recognizable even though the model
is fit well.

\newcommand{\mnistwidth}{0.6in}

\begin{figure}[h]
\centering{\subfigure[]{
\includegraphics[width=\mnistwidth]{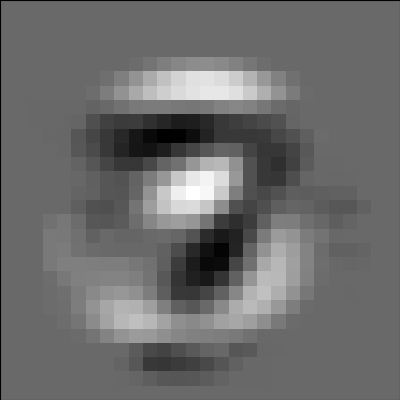}
\label{fig:lin_32b}}} 
\centering{\subfigure[]{
\includegraphics[width=\mnistwidth]{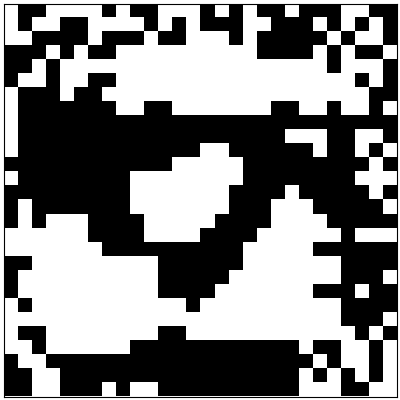}
\label{fig:lin_1b}}} 
\centering{\subfigure[]{
\includegraphics[width=\mnistwidth]{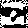}
\label{fig:logreg_weights_sign}}} 
\centering{\subfigure[]{
\includegraphics[width=\mnistwidth]{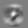}
\label{fig:logreg_weights}}} 
\caption{Visualization of~\subref{fig:lin_32b} logistic regression model
weights after expert initialization, and fine-tuning for 10k steps on the
sigmoid cross entropy loss with Adam~\citep{kingma2014adam}, a batch size of
$128$, learning rate of 1e-5, and an overall $L_2$ regularization
constant of $0.05$.~\subref{fig:lin_1b} A binarized set of weights
after training with the same procedure from
scratch for 50k steps. Plots~\subref{fig:logreg_weights_sign}
and~\subref{fig:logreg_weights} are taken from~\cite{goodfellow2015explaining}
for comparison purposes,~\subref{fig:logreg_weights_sign} is an FGSM
perturbation for true class label ``7'',
and~\subref{fig:logreg_weights} the actual weights. The faint
dark pixels in~\subref{fig:logreg_weights} resemble a slightly rotated
``?'' symbol, wheras ``3'' and ``7'' are visible in~\subref{fig:lin_32b}.}
\label{fig:mnist_three_seven}
\end{figure}

We argue that there~\emph{is} a clear interpretation of such a perturbation.
In this particular case, one's intuition is partly obscured due to having
used the sign of the gradient (FGSM) rather than the real
gradient. Although the $L_\infty$ bounded FGSM~\emph{is} the strongest attack
against the logistic regression model, it
is not the most~\emph{efficient} attack. The gradient sign in
Figure~\ref{fig:logreg_weights_sign} is recognizable as binarizing the
continuous valued weights, and the roughly 3--5 pixel wide noisy
border reflects directions where the weights are close to, but not exactly
zero\footnote{Implementations of the~\texttt{sign} method, e.g.
in~\texttt{numpy} and TensorFlow do pass through $0$ as $0$.}. We can induce a
nearly identical error rate, but with less distortion by using the actual
gradient, and normalizing it to have unit $L_2$ norm, the ``Fast grad $L_2$''
method from~\cite{kurakin2017adversarial} (their Appendix A).

The perturbation image shown in Figure~\ref{fig:logreg_weights_sign} for the
FGSM attack is approximately identical to the binary weights
of a binarized model trained on this task from scratch, as shown in
Figure~\ref{fig:lin_1b}.
We know the binarized model is~\emph{not} an optimal model here, as we
would like it to be totally invariant to the areas around the border where the
real-valued model's weights, and the difference between the
average ``7'' and ``3'', are almost zero. The binarized model tries to
be as invariant as possible to the background by roughly equally distributing
the polarity of weights in these areas.  Cropping the dataset more closely
around the digits, or allowing a third parameter state (e.g.\ by using a ternary
network, or pruning connections to bordering pixels) would make the model more
robust. In terms of the boundary tilting perspective, such a
model~\emph{minimizes tilting along dimensions of
low-variance in the data}~\citep{tanay2016boundary}.

Using the expert initialization scheme and training procedure from
Figure~\ref{fig:lin_32b}, our ``Expert-$L_2$'' model achieves $59\%$ test error
for an FGSM attack, compared to the $99\%$ baseline error rate,
for the same perturbation size $\epsilon=0.25$ w.r.t.~the $L_\infty$ norm.
This 40\% improvement is consistently reproducible, but the model
incurs a $4\%$ error rate on the natural test set, compared to $1.6\%$ error
for the baseline model. We believe the $2.4\%$ loss in ``clean'' test accuracy
in exchange for a $40\%$ gain for this attack is a good trade-off.
After a brief search, we find that a randomly initialized model can handle very
large norm penalties. For example, training from scratch for 50k steps with
$\lambda = 3.25$ maintains the $\approx 4\%$ test error of the Expert-$L_2$
model, while achieving $43\%$ and $64\%$ FGSM error rates for $L_1$ and $L_2$
weight decay respectively.
Next we explain why we prefer penalizing the $L_2$ norm of the weights
rather than the $L_1$ norm, despite the slightly higher FGSM error rate
for the former.

It may seem that a $59\%$ FGSM error for $\epsilon=0.25$
is still unreasonably high, but we argue that this data point is fairly
uninformative in isolation and does not imply robustness, or a lack thereof.
With exceedingly high $L_1$ weight decay (e.g.~with $\lambda=32$), we can
further reduce the FGSM error rate to $25\%$, but what do we suspect the
robustness behaviour of this model will be?

Even for the linear case where the FGM is exact, one score does not tell the
full story. Although these types of comparisons are convenient for researchers,
they are potentially harmful in the long run as this rewards overfitting a
particular threat model, which we have already observed. We leave it up to
researchers to rigorously attack their proposed defenses, and suggest that
characterizing the slope of the accuracy roll-off curve for a variety of
optimization-based or engineered attacks makes a more compelling defense. As a
further assessment, the nature of the distortion introduced by the attack may
be of interest when expressed quantitatively and qualitatively. The $59\%$
error rate for Expert-$L_2$ coincides with a mean $L_2$ distortion of $\approx
4\%$ over the full test set. As this distortion metric is not
always representative of perceptual distance (e.g.~it is not translation
invariant), we include candidate adversarial images for Expert-$L_2$
in Figure~\ref{fig:linear_mnist_adv_ex}.

\newcommand{\exwidth}{0.35in}

\begin{figure}[h]
\centering{\subfigure[]{
\includegraphics[width=\exwidth]{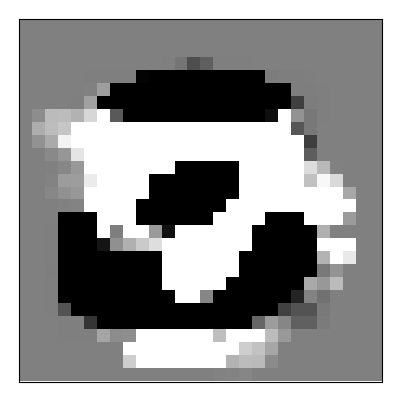}
\label{fig:linear_grads}}} 
\centering{\subfigure[]{
\includegraphics[width=\exwidth]{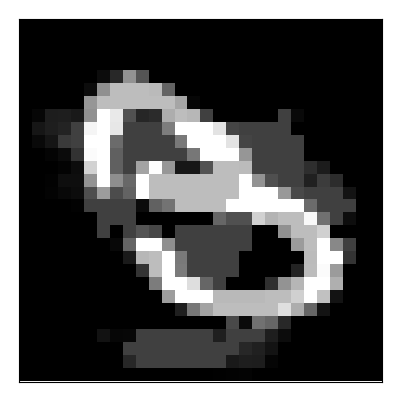}
\label{fig:adv_3_eps_0p25_s0}}} 
\centering{\subfigure[]{
\includegraphics[width=\exwidth]{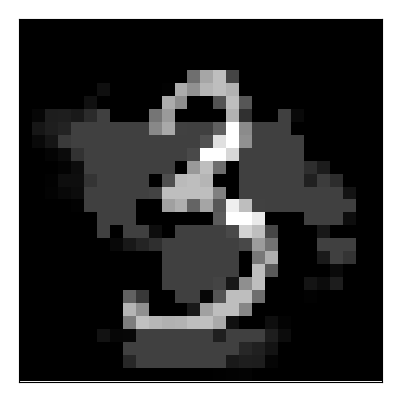}
\label{fig:adv_3_eps_0p25_s1}}} 
\centering{\subfigure[]{
\includegraphics[width=\exwidth]{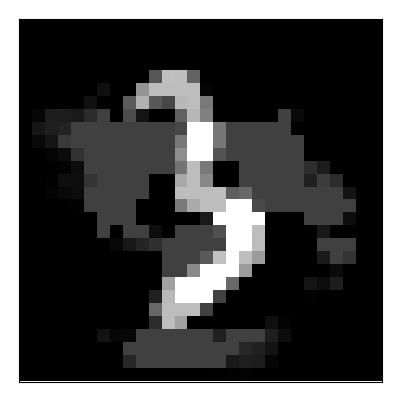}
\label{fig:adv_3_eps_0p25_s2}}} 
\centering{\subfigure[]{
\includegraphics[width=\exwidth]{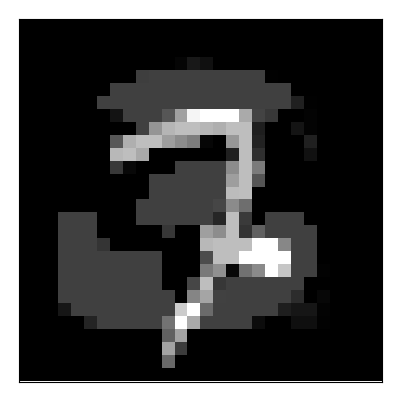}
\label{fig:adv_7_eps_0p25_s0}}} 
\centering{\subfigure[]{
\includegraphics[width=\exwidth]{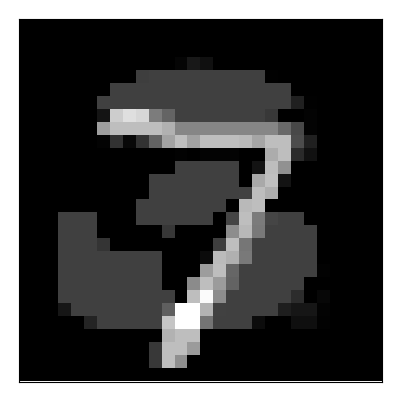}
\label{fig:adv_7_eps_0p25_s1}}} 
\centering{\subfigure[]{
\includegraphics[width=\exwidth]{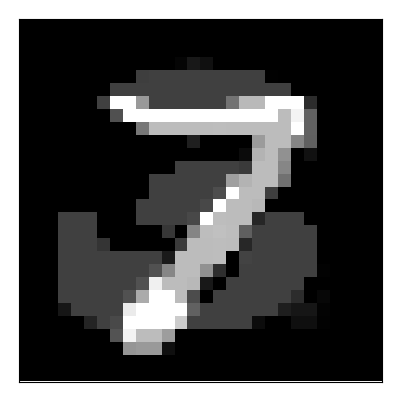}
\label{fig:adv_7_eps_0p25_s2}}} 
\caption{\subref{fig:linear_grads} Fast grad $L_2$ perturbation
for original class ``3'', and candidate adversarial examples
for original class
``3''~[\subref{fig:adv_3_eps_0p25_s0}--\subref{fig:adv_3_eps_0p25_s2}]
and ``7''~[\subref{fig:adv_7_eps_0p25_s0}--\subref{fig:adv_7_eps_0p25_s2}].
All images remain correctly classified when perturbation bounded
by $\epsilon=0.25$ in the $L_\infty$ norm, we therefore call
them only~\emph{candidate} examples as in~\cite{kurakin2017adversarial}.
The average ``3'' is clearly visible in images originally
depicting a ``7'', and vice versa.}
\label{fig:linear_mnist_adv_ex}
\end{figure}

To show that an attack causing less distortion than FGSM can still be
interpretable, we adapt the Fast grad $L_2$ method by scaling the
normalized gradient by $100\times$ before clipping to $\pm \epsilon$.
Scaling by 100 lets us approximate the higher error rate caused by
FGSM, while minimizing unnecessary distortion. This attack results in an error
rate of $58.9\%$, compared to $59.1\%$ on our Expert-$L_2$ model when vanilla
FGSM is used, but causes
$4.00\%$--$L_2$ distortion compared to $4.57\%$ for FGSM\@. Clearly the
images in Figure~\ref{fig:linear_mnist_adv_ex} are ambiguous even for human
vision, and correct classification could be contingent on having previously
seen a sampling of natural examples. These examples support the observation
of~\cite{tanay2016boundary} that:

\begin{quotation}
\dots the weight vector found by logistic regression points in a direction that
is close to passing through the mean images of the two classes, thus defining a
decision boundary similar to the one of a nearest centroid classifier.
\end{quotation}

\section{Adversarial Training Versus Weight Decay for Linear Models}
\label{sec:adv-vs-wd-linear-models}

In linear classification, weight decay is often explained as a way of enforcing
a preference for models with small weights (e.g.\ section 5.2.2 of
\citep{goodfellow2016deep}). Yet there is no reason, \emph{a priori}, for
models with small weights to generalize better: all the weight vectors
$\,\lambda\, w\,$ with $\lambda > 0$ define the same hyperplane boundary,
whether $\lambda$ is small or large. The norm of the weight vector is only
meaningful in the context of ERM, because it influences the scaling of the loss
function~\citep{neyshabur2017exploring,tsuzuku2018lipschitz}.

We first introduce this natural interpretation of weight decay as affecting the
scaling of the loss function, and then directly compare variants of adversarial
training to $L_1$ and $L_2$ weight decay.

\subsection{Theoretical Analysis}
\label{sec:theoretical-analysis}

Consider a logistic regression model with input $x \in \mathbb{R}^d$,
labels $y\in\{\pm1\}$, parameterized by weights $w$, bias $b$, and having
sigmoid non-linearity $\sigma(z)$. Predictions are defined
by $\sigma(w^\top x + b)$ and the model can be optimized by stochastic
gradient descent (SGD) on the sigmoid cross entropy loss. This reduces to
SGD on~\eqref{eq:objective1} where $\zeta$ is the softplus loss\footnote{Our
function $\zeta$ is the mirror of the one used
in~\cite{goodfellow2015explaining}:
$\zeta(z) \leftarrow \zeta(-z)$. This definition removes the superfluous
negative sign inside $\zeta$ later on.} $\zeta(z) = \log(1 + e^{-z})$:
\begin{equation}
\mathbb{E}_{x, y \sim p_\text{data}}\; \zeta(y (w^\top x + b))
\label{eq:objective1}
\end{equation}

In Figure~\ref{fig:wd_vs_at1}, we develop intuition for the different quantities
contained in~\eqref{eq:objective1} with respect to a typical binary
classification problem. We will later see how this analysis generalizes to the
multi-class setting.

\newcommand{\cellwidth}{0.95in}

\begin{figure}[h]
\begin{center}
\centering{\subfigure[$w^\top x + b$]{
\includegraphics[width=\cellwidth]{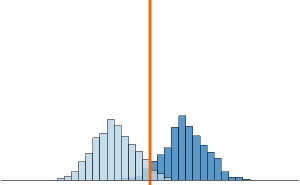}
\label{fig:wd_vs_at_cell1}}} 
\centering{\subfigure[$y (w^\top x + b)$]{
\includegraphics[width=\cellwidth]{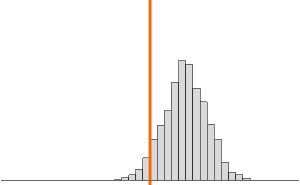}
\label{fig:wd_vs_at_cell2}}} 
\centering{\subfigure[$\zeta(y (w^\top x + b))$]{
\includegraphics[width=\cellwidth]{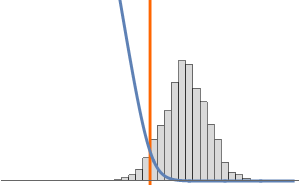}
\label{fig:wd_vs_at_cell3}}} 
\end{center}
\caption{\subref{fig:wd_vs_at_cell1} For a given weight vector $w$ and bias $b$,
the values of $\,w^\top x + b\,$ over the training set typically follow a
bimodal distribution (corresponding to the two classes) centered on the
classification boundary. \subref{fig:wd_vs_at_cell2}  Multiplying by the label
$y$ allows us to distinguish the correctly classified data in the positive
region from misclassified data in the negative
region.~\subref{fig:wd_vs_at_cell3} We can then attribute a penalty to each
training point by applying the softplus loss to $y(w^\top x + b)$.}
\label{fig:wd_vs_at1}
\end{figure}

We note that $\;w^\top x + b\;$ is a \emph{scaled}, signed distance between $x$
and the classification boundary defined by our model. If we define $d(x)$ as the
signed Euclidean distance between $x$ and the boundary,
then we have: $w^\top x + b = \|w\|_2\,d(x)$. Hence,
minimizing~\eqref{eq:objective1} is equivalent to minimizing:
\begin{equation}
\mathbb{E}_{x, y \sim p_\text{data}}\; \zeta(\|w\|_2 \times y\,d(x))
\label{eq:objective2}
\end{equation}
We define the \emph{scaled softplus loss} as:
\begin{equation}
\zeta_{\|w\|_2}(z) := \zeta(\|w\|_2 \times z)
\label{eq:scaled-softplus}
\end{equation}
And we see that adding a $L_2$ weight decay term in~\eqref{eq:objective2},
resulting in~\eqref{eq:objective3}, can be understood as a way of controlling the
scaling of the softplus function:
\begin{equation}
\mathbb{E}_{x, y \sim p_\text{data}}\; \zeta_{\|w\|_2}(y\,d(x)) + \lambda\|w\|_2
\label{eq:objective3}
\end{equation}

\begin{figure}[h]
\begin{center}
\centering{\subfigure[$\zeta_{5}(y\,d(x))$]{
\includegraphics[width=\cellwidth]{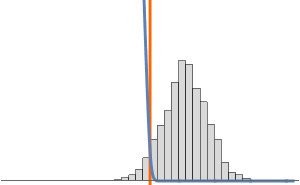}
\label{fig:wd_vs_at_cell4}}} 
\centering{\subfigure[$\zeta_{0.5}(y\,d(x))$]{
\includegraphics[width=\cellwidth]{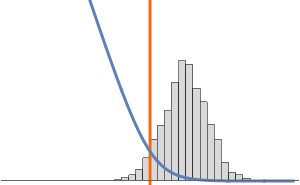}
\label{fig:wd_vs_at_cell5}}} 
\centering{\subfigure[$\zeta_{0.05}(y\,d(x))$]{
\includegraphics[width=\cellwidth]{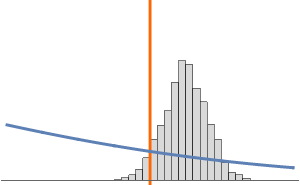}
\label{fig:wd_vs_at_cell6}}} 
\end{center}
\caption{\subref{fig:wd_vs_at_cell4} For a small regularization parameter
(large $\|w\|_2$), the misclassified data is penalized linearly while the
correctly classified data is not penalized.~\subref{fig:wd_vs_at_cell5} A
medium regularization parameter (medium $\|w\|_2$) corresponds to smoothly
blending the margin.~\subref{fig:wd_vs_at_cell6} For a large regularization
parameter (small $\|w\|_2$), the entire data is penalized almost linearly.}
\label{fig:wd_vs_at2}
\end{figure}

Figure~\ref{fig:wd_vs_at2} illustrates the effect of varying $\lambda$, the
regularization parameter for $\|w\|_2$.

\begin{figure}[h]
\begin{center}
\centering{\subfigure[$\zeta_{1}(y\,d(x))$]{
\includegraphics[width=\cellwidth]{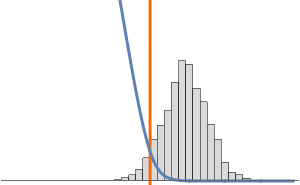}
\label{fig:wd_vs_at_cell7}}} 
\centering{\subfigure[$\zeta_{1}(y\,d(x)-3)$]{
\includegraphics[width=\cellwidth]{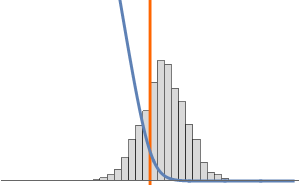}
\label{fig:wd_vs_at_cell8}}} 
\centering{\subfigure[$\zeta_{1}(y\,d(x)-13)$]{
\includegraphics[width=\cellwidth]{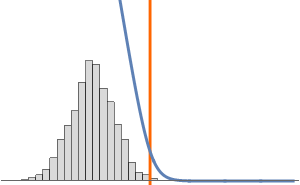}
\label{fig:wd_vs_at_cell9}}} 
\end{center}
\caption{\subref{fig:wd_vs_at_cell7} No adversarial training.
\subref{fig:wd_vs_at_cell8} Moderate adversarial training effectively
enforces a larger margin by translating the data into the misclassified
region, while softplus scaling remains unaffected.~\subref{fig:wd_vs_at_cell9}
With strong adversarial training, all the data is in the misclassified region.
This regime is unstable and we expect models with limited statistical
complexity to fail to converge, which we observe experimentally.}
\label{fig:wd_vs_at3}
\end{figure}

Similarly, we can interpret adversarial training as acting on the horizontal
offset of $y\,d(x)$. Training on worst case adversarial examples consists of
replacing $x$ with $x^\prime = x - \epsilon\,y\,\hat{w}$ (where $\hat{w} =
w/\|w\|_2$). We then have:
\begin{align*}
y\,(w^\top x^\prime + b) &\,=\, y\,(w^\top (x - \epsilon\,y\,\hat{w}) + b)\\
												 &\,=\, y\,(w^\top x  + b) - \epsilon\,\|w\|_2\\
												 &\,=\, \|w\|_2 \times (y\,d(x) - \epsilon)
\end{align*}
Substituting the above expression into the scaled softplus
function~\eqref{eq:scaled-softplus} yields:
\begin{equation}
\mathbb{E}_{x, y \sim p_\text{data}}\; \zeta_{\|w\|_2}(y\,d(x) - \epsilon)
\label{eq:objective4}
\end{equation}
Figure~\ref{fig:wd_vs_at3} illustrates the effect of varying the adversarial
perturbation magnitude, or offset parameter $\epsilon$.

The softplus function has a linear region penalizing the misclassified data,
a saturated region leaving the correctly classified data unaffected, and a
smooth transition between these two regions enforcing a safety margin. Both
weight decay and adversarial training can be interpreted as influencing how
these different regions interact with the data.
Weight decay strikes an effective balance between the two extreme regimes:
when it is very low, the margin blend is very sharp, as seen in
Figure~\ref{fig:wd_vs_at_cell4}. When it is very high, nearly all of the data
falls into the margin, shown in Figure~\ref{fig:wd_vs_at_cell6}, resulting in
underfitting, but no risk of instability as the data does not shift.
Adversarial training artificially increases the size of the margin by
translating some or all of the data into the misclassified region, as in
Figure~\ref{fig:wd_vs_at_cell8}.
It becomes unstable when the offset used is too strong, which we depict in
Figure~\ref{fig:wd_vs_at_cell9} and verify experimentally in
Section~\ref{sec:adv-train-vs-wd}. Overall, weight decay seems more
theoretically sound given that it allows a broader variety of behaviors and
remains stable for~\emph{all} regimes, whereas the adversarial training
framework is inherently limited to small perturbations. This translation and
scaling perspective explains why adversarial training and weight decay can
complement each other when small perturbations are used, but ultimately, weight
decay is the more sound instrument.

In this section we focused on the $L_2$ norm case which is, in our opinion,
more intuitive since $d(x)$ can then be interpreted as the Euclidean distance
from the boundary. However, $L_1$ weight decay and FGSM adversarial training
can be shown to implement similar transformations in a different normed space,
thus enforcing robustness to different types of perturbations. We perform
experiments using both norms in the following section.

\subsection{Experiments}
\label{sec:adv-train-vs-wd}


Consider the same logistic regression model as in previous sections. We use
the standard form of the softplus loss\footnote{with positive $z$ this time.}
where $\zeta(z)=\log(1 + e^z)$ to facilitate a direct comparison
with~\citep{goodfellow2015explaining}. Training the model consists in minimizing
the loss~\eqref{eq:std-objective} with SGD\@:
\begin{equation}
\mathbb{E}_{x, y \sim p_\text{data}} \zeta( -y (w^\top x + b)).
\label{eq:std-objective}
\end{equation}
The gradient of~\eqref{eq:std-objective} with respect to the input $x$ is $-yw$,
and its sign is $-y\sign(w)$. Note that the label $y$ was
omitted in the original~\citeauthor{goodfellow2015explaining}\ paper, resulting
in an asymmetric penalty for adversarial training. Training on worst case
$L_\infty$ bounded examples for logistic regression consists of replacing $x$
in~\eqref{eq:std-objective} with $x - \epsilon y \sign(w)$, resulting
in~\eqref{eq:adv-train}.
The perturbation term $\epsilon y w^\top\sign(w)$ is equivalent to an $\epsilon
y \|w\|_1$ penalty on the weights.
\begin{equation}
\mathbb{E}_{x, y \sim p_\text{data}} \zeta(-(y (w^\top x + b) - \epsilon \|w\|_1))
\label{eq:adv-train}
\end{equation}
Compare the expression~\eqref{eq:adv-train} to instead adding the $L_1$ penalty
controlled by regularization factor $\lambda$, outside of $\zeta(z)$ as
in~\eqref{eq:train-wd}:
\begin{equation}
\mathbb{E}_{x, y \sim p_\text{data}} \zeta( -y (w^\top x + b)) + \lambda\|w\|_1.
\label{eq:train-wd}
\end{equation}
We now provide a learning theoretic observation of the differences
between the two expressions, which complements the scaling and translation
view. The parameter norm penalty in~\eqref{eq:adv-train} is trapped in $\zeta(z)$, and
interacts with $y$ (the true or predicted label), but this
is not the case with weight decay~\eqref{eq:train-wd}. This leads to a possible
perverse instantiation of the ERM minimization objective
in~\eqref{eq:adv-train} if $\epsilon \|w\|_1$ overpowers $-(w^\top x + b)$, as
the model may intentionally learn to make wrong predictions in order
to ``escape'' the saturating region of $\zeta$ and maximize weight updates
for shrinking $\epsilon \|w\|_1$. Although~\eqref{eq:train-wd} also causes
the model to trade-off accuracy on the training data in order to keep
$\lambda \|w\|_1$ small, there is no risk of accidentally teaching
a~\emph{preference} for false predictions on the natural data
for very large $\lambda$.

\citeauthor{goodfellow2015explaining}~suggested that a possible benefit
of adversarial training is that the $L_1$ penalty is \emph{subtracted} from
the activation, rather than~\emph{added} to the training cost, therefore the
penalty could disappear if the model makes predictions confident enough to
saturate $\zeta$. However, this is only true for confident~\emph{correct}
predictions, which are only possible for small $\epsilon$.
Otherwise $\zeta$ is operated in the non-saturating linear region.
More importantly, the translation effect artificially increases the margin
between classes at the risk of overfitting, whereas weight decay
encourages underfitting.
They did not report any experiments where weight decay was applied to
the~\emph{logistic regression} model currently under consideration, and only
allude to a brief experiment where small (e.g.\ $\lambda=0.0025$) $L_1$
penalties were used on the first layer of a maxout
network~\citep{goodfellow2013maxout} with an unreported architecture and depth.
It is also unclear if this was done independently, or concurrently, with
adversarial training, and for how many steps or epochs. Ultimately, only one
data-point for one trial was provided, a 5\% error on the training set.

In accordance with the theoretical discussion of equations~\eqref{eq:adv-train}
and~\eqref{eq:train-wd}, we observe a positive feedback loop
when using FGSM training, and find that it does not deactivate
automatically, as suggested it may when it was originally proposed.
Although weight decay also does not deactivate automatically per se,
we argue that this is not a limitation provided that one carefully tunes
$\lambda$ to obtain a suitable scaling of the loss with respect to the number
of examples and variance in the data.

Adversarial training with medium $\epsilon$ (e.g.\ $0.25$) proceeds as follows:
initially a randomly initialized model poorly fits the training data and
$\zeta$ is in the linear region, staying there for much of the overall training
time. In the linear region of $\zeta$, it is as if the $L_1$ penalty is added
to the cost, as in~\eqref{eq:train-wd}.
\[
\mathbb{E}_{x, y \sim p_\text{data}} \zeta( -y (w^\top x + b)) + \epsilon\|w\|_1
\label{eq:y-out}
\]
Predictions of either class are systematically reduced
by $\epsilon\|w\|_1$, which leads to the bimodal distribution shown
in~Figure~\ref{fig:wd_vs_at_cell1} shifting inward. Activations, and
therefore $\|w\|_1$,~\emph{decrease} monotonically, while the
training loss~\emph{increases} monotonically since predictions are made with
less confidence. When we finally enter the non-linear region of $\zeta$, the
training loss is high, but there are few non-zero parameters left to update,
resulting in aggressive changes to just a few weights.
Beyond a certain point, not even strong weight decay can rescue learning if we
consider training on the bi-stable examples generated with a constant $\epsilon$
in Figure~\ref{fig:linear_mnist_adv_ex}. In the case of a near~\emph{perfect}
decision boundary, approximately that defined by the weights visualized in
Figure~\ref{fig:lin_32b}, and $\epsilon$ larger than the distance from an
example to the boundary, we get examples that literally belong to
the~\emph{opposite} class. Adversarial training becomes even more destructive,
the closer a model is toward converging near a suitable solution.

We acknowledge that in most practical applications of interest we don't have
a perfect boundary, but this simple case clearly contradicts
the notion that the penalty disappears on its own,~\emph{at least in the way
that we would like it to}. There is no reason to expect the situation will
improve for higher-dimensional higher-variance problems like CIFAR-10 or
ImageNet.

\newcommand{\fgsmwidth}{0.4in}

\begin{figure}[h]
\centering{\subfigure[]{
\includegraphics[width=\fgsmwidth]{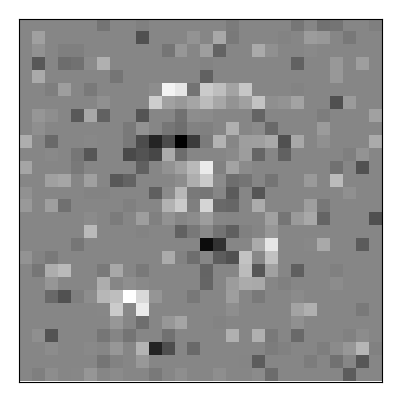}
\label{fig:linear_fgsm_train_10k}}} 
\centering{\subfigure[]{
\includegraphics[width=\fgsmwidth]{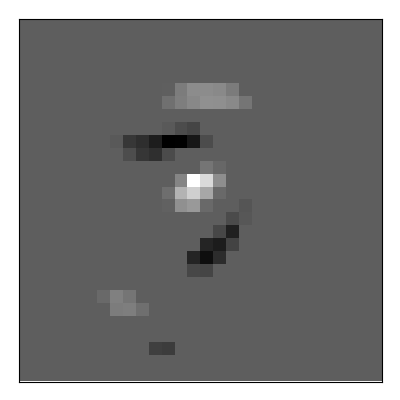}
\label{fig:linear_fgsm_train_50k}}} 
\centering{\subfigure[]{
\includegraphics[width=\fgsmwidth]{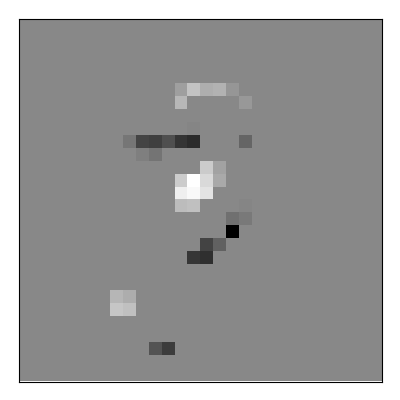}
\label{fig:linear_50kscratch_l1_15}}} 
\centering{\subfigure[]{
\includegraphics[width=\fgsmwidth]{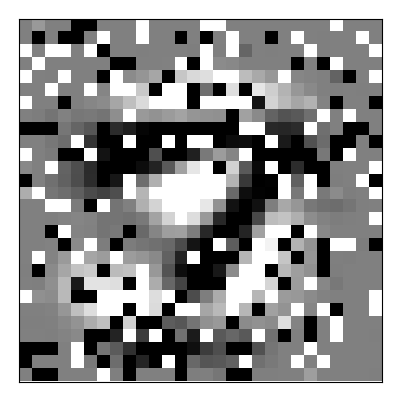}
\label{fig:linear_fast_grad_10kscratch_pert}}} 
\centering{\subfigure[]{
\includegraphics[width=\fgsmwidth]{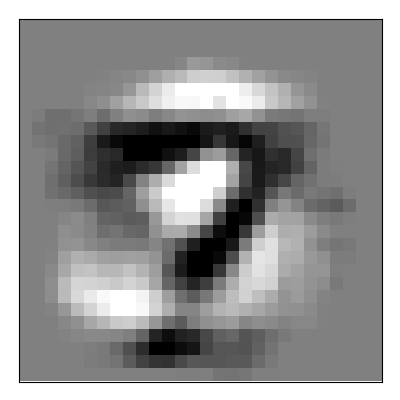}
\label{fig:linear_fast_grad_50kscratch_pert}}} 
\centering{\subfigure[]{
\includegraphics[width=\fgsmwidth]{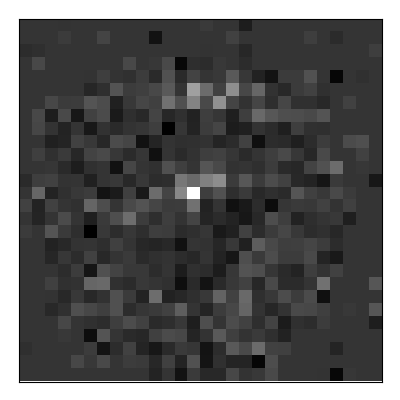}
\label{fig:linear_fast_grad_50kscratch}}} 
\caption{Weight visualization for~\subref{fig:linear_fgsm_train_10k}
$\epsilon=0.25$ FGSM training for 10k steps,
and~\subref{fig:linear_fgsm_train_50k} 50k
steps.~\subref{fig:linear_50kscratch_l1_15} Training instead for 50k steps only
with $L_1$ weight decay and $\lambda = 15$. Training with the same $\epsilon$
for Fast grad $L_2$ results in similar weights (not shown) as
in~\subref{fig:linear_fgsm_train_10k} and~\subref{fig:linear_fgsm_train_50k},
but a more interpretable perturbation image after
10k~\subref{fig:linear_fast_grad_10kscratch_pert}, and 50k
steps~\subref{fig:linear_fast_grad_50kscratch_pert}. Increasing $\epsilon$ to
$0.5$ results in non-stability and fails to train as the model lacks sufficient
capacity to overfit the residual noise arising from overlapping the class
centroids~\subref{fig:linear_fast_grad_50kscratch}.}
\label{fig:linear_fgsm_training}
\end{figure}

Figure~\ref{fig:linear_fgsm_training} shows the effect of adversarial training
on our logistic regression model. FGSM adversarial training always converges to
the solution in Figure~\ref{fig:linear_fgsm_train_50k} given a sufficient
number of steps, regardless of whether the~\emph{true label} or the model's own
~\emph{prediction} is used to construct the adversarial examples,
independent of expert initialization or otherwise, and of
weight decay being used or not. In other words, FGSM adversarial training
with $\epsilon=0.25$ always overpowers our initialization scheme and the $L_2$
weight decay with $\lambda = 0.05$, converging on this solution that obtains
seemingly good accuracy for two benchmarks, around 96\% for the clean test
set and 71\% for FGSM with $\epsilon = 0.25$. However, it should be
apparent on inspection of the weights that this model is not a good fit for
this data.
Notice in Figure~\ref{fig:linear_fast_grad_10kscratch_pert} that training with
Fast grad $L_2$ and medium $\epsilon$ results in a low-frequency perturbation
image like the weights of Expert-$L_2$, but contains high-frequency artifacts
due to the sharp transition between the linear and saturating region of the
softplus loss in this regime due to the apparent translation of the data.
When $\epsilon$ further increases in
Figure~\ref{fig:linear_fast_grad_50kscratch}, there is nothing left to fit but
noise.

These degenerate solutions that obtain high FGM accuracy fail
the~\emph{fooling images} test as they cannot generate natural looking data
examples when gradient ascent is performed on the logit(s), despite making such 
predictions with high-confidence~\citep{nguyen2015deep,
galloway2018predicting}. Additionally, an $L_0$ adversary can
quickly cause a model tuned for the $L_\infty$ metric to predict at no
better than chance by setting a small number of pixels corresponding to the
locations of the min and max weights in Figures~\ref{fig:linear_fgsm_train_50k}
and~\ref{fig:linear_50kscratch_l1_15} to~1\@. The effect of this black-box $L_0$
attack is depicted in Figure~\ref{fig:logistic_pixel_attack}, and shows how
adversarial training can behave like the classic ``whack-a-mole'' game. Even
though we used expert knowledge to craft this attack, it is unreasonable to
assume that attackers would not be able uncover similar information given
enough time. The attack also simulates the effect of ``dead pixels'' which
could arise in a robotic system or sensor deployed in the wild.
\newcommand{\attkwidth}{1.46in}
\begin{figure}[h]
\centering{\subfigure[]{
\includegraphics[width=\attkwidth]{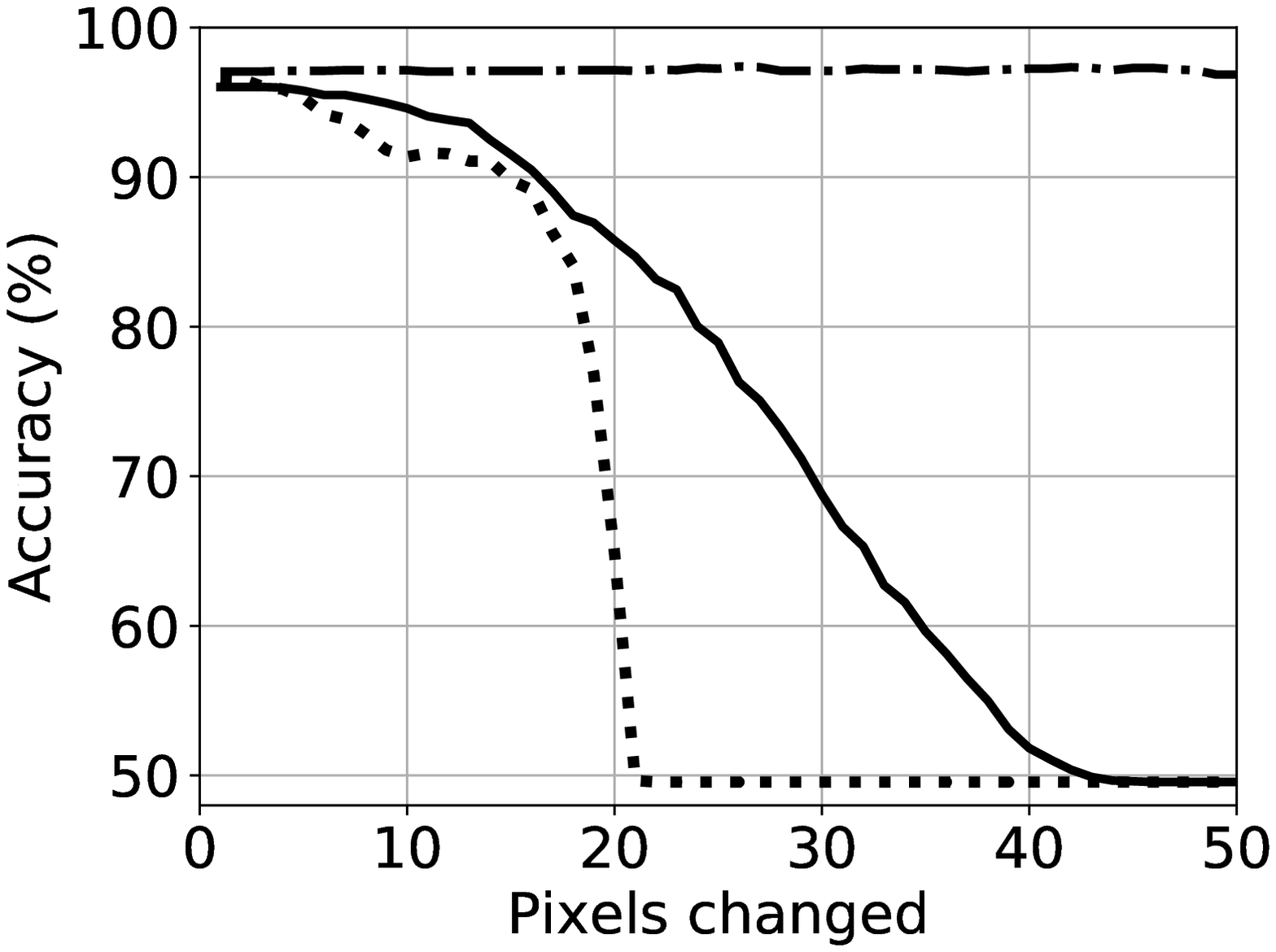}
\label{fig:logistic_pixel_attack}}} 
\centering{\subfigure[]{
\includegraphics[width=\attkwidth]{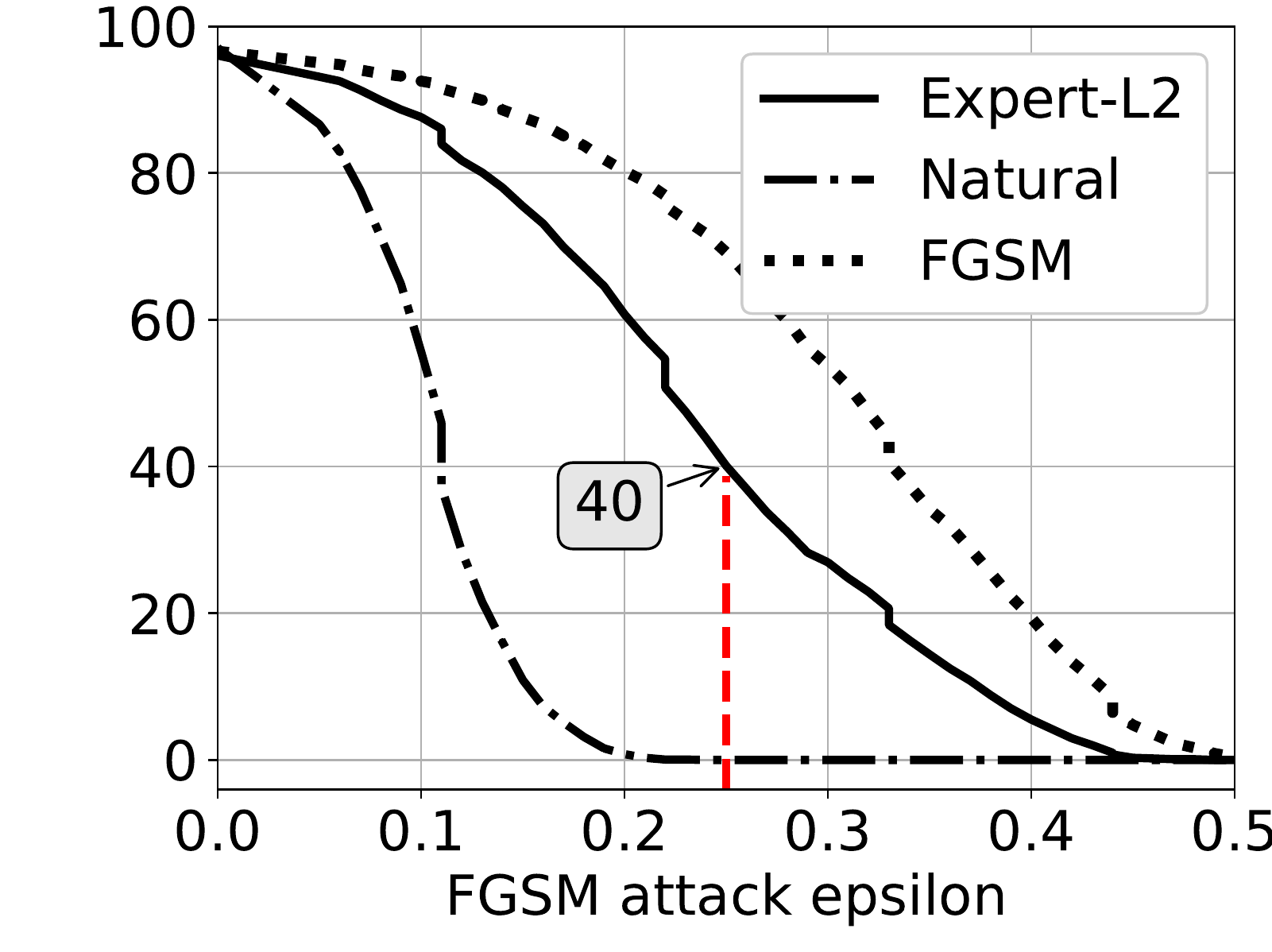}
\label{fig:logistic_fgsm_attack}}} 
\caption{A comparison between~\subref{fig:logistic_pixel_attack} an $L_0$
attack, and~\subref{fig:logistic_fgsm_attack} $L_\infty$--FGSM\@.~Models
trained with FGSM were more robust in~\subref{fig:logistic_fgsm_attack},
but less so in~\subref{fig:logistic_pixel_attack}.~An undefended model
(``Natural'') was robust in~\subref{fig:logistic_pixel_attack}, but
not~\subref{fig:logistic_fgsm_attack}.~Expert-$L_2$ was reasonably robust
to~\emph{both}.~The model trained
on FGSM examples used $\epsilon=0.25$, and the $x$-intercept
at $\approx 20$ pixels in~\subref{fig:logistic_pixel_attack}
is translated left and right by training with different $\epsilon$.
Annotated is Expert-$L_2$ accuracy corresponding to
the $\epsilon=0.25$--FGSM error rate discussed
in Section~\ref{sec:robust-logistic}.}
\label{fig:logistic_attacks}
\end{figure}

Although the linear model has limited capacity, it is surprisingly robust to
an FGSM attack when trained with FGSM, as shown in
Figure~\ref{fig:logistic_fgsm_attack}. This is at odds with the recommendation
of~\citeauthor{madry2018towards}, whom suggest increasing capacity to train
with a strong adversary. We argue that this becomes an arms race to use as much
capacity as possible, and strongly advise against this as it encourages
poor generalization outside the considered threat model.
In this case, for the $L_\infty$ metric FGSM is the strongest possible
adversary, and the model appears robust until we consider our trivial $L_0$
attack.
Note that there was an insignificant difference between using the true label or
the model's own prediction when generating the training examples, since
predictions closely matched the true labels. To summarize the results of
Figure~\ref{fig:logistic_attacks}: models trained with FGSM were robust to
FGSM, but not to an $L_0$ attack. Undefended models were robust to the $L_0$
attack, but not FGSM\@. Models trained with $L_2$ weight decay were
reasonably robust to~\emph{both}, and their performance declined slowly and
predictably for unbounded perturbations.

\subsection{Discussion}

When a randomly initialized model is trained without weight
decay, spurious data are fit more tightly, leaving the boundary near the
data sub-manifolds. Adversarial training examples generated with such a
boundary are projected by only a small distance, having a small but
increasingly positive effect. Given that the effect is small initially,
there is a natural inclination to use a large $\epsilon$. Medium to high
$\epsilon$ adversarial training seems to work well in practice for
high-capacity deep neural networks because they are able to fit the
signal~\emph{and} the noise, an effect we aren't the first to
observe~\citep{zhang2017understanding}.

To use adversarial training in a more principled manner implies that $\epsilon$
should at the very least be decayed when training, just as is commonly done
with learning rates. Ultimately, the value of $\epsilon$ used with adversarial
training is a poorly characterized constant that has been selected with limited
intuition in much of the literature to date. To be mathematically correct,
$\epsilon$ should be a function of each example's distance from its class
mean. Yet another hyper-parameter is introduced, as we must set both an
$\epsilon$ decay schedule, as well as a unique $\epsilon$ per each image. This
brings us back to weight decay, which typically only has one hyper-parameter,
the regularization constant $\lambda$, and whose behaviour is reasonably well
characterized at several levels of abstraction. We say~\emph{reasonably} well
because its exact behaviour remains somewhat of a mystery in deep
learning~\citep{zhang2017understanding}, and if it was fully characterized then
it would have been used in the baseline work to defend their logistic
regression model against imperceptible adversarial examples.

There is ample room to further understand how norm constraints
compare to other types of explicit and
implicit regularizers used in deep learning,
and how to bound generalization error in this
setting~\citep{neyshabur2017exploring}. As we will see in
Section~\ref{sec:cifar-exp}, it can be surprising that $L_2$ weight
decay has a filter-level pruning effect when used with CNNs, rather than
simply making all elements~\emph{small}.~What remains in the first
layer are smooth filters that implement fundamental image processing
primitives, somewhat similar to the fully-connected weights of our Expert-$L_2$
model.

\section{Deep Neural Networks}


It was suggested in the seminal work of~\cite{szegedy2014intriguing} that
the existence of adversarial examples in neural networks is related to the
notion of Lipschitz constant, which measures the least-upper-bound of the
ratio between output distances and corresponding input distances. This idea
was further developed and formulated explicitly by~\cite{cisse2017parseval},
who suggested that a neural network's sensitivity to adversarial
perturbations can be~\emph{controlled} by the Lipschitz constant.
Yet this argument is incomplete: just like rescaling the weight vector of a
linear classifier does not affect the hyperplane boundary it defines,
varying the Lipschitz constant of a neural network by rescaling its
output affects neither its predictions nor its vulnerability to
adversarial perturbations~\citep{tsuzuku2018lipschitz}.
Instead, we argue that the Lispchitz constant is only meaningful in
the context of ERM because it influences the way the loss
function distributes penalties over the training data
\citep{neyshabur2017exploring,tsuzuku2018lipschitz}.
Below, we generalize our considerations of
Section~\ref{sec:theoretical-analysis}
on binary linear classification to multiclass neural networks and we then
reinterpret weight decay and adversarial training as part of a broader
family of~\emph{output regularizers}.

\subsection{Generalizing the Binary Linear Case}

Consider a feedforward network $\mathcal{N}$ consisting of $l$
layers with parameters $W_1, \dots, W_l$ and Rectified Linear Unit activations
$\phi$. We assume that $\mathcal{N}$ does not use any weight or batch
normalization and for simplicity, we also neglect the role of the biases.
For an input image $x$, the logit layer of $\mathcal{N}$ can be written:
$$z(x) = W_l\,\phi(W_{l-1}\,\phi(\dots\,\phi(W_1\,x)))\;	.$$
For a scalar $\lambda$ we have $\phi(\lambda\, x) = \lambda\,\phi(x)$ and
therefore, decaying all the weights $W_i \leftarrow \lambda\,W_i$ is
equivalent to decaying the logits:
$$z(x) \leftarrow \lambda^l\,z(x).$$
In binary classification, $z$ is a scalar and $\lambda^l$ acts as a
scaling parameter for the softplus function. In multiclass classification, $z$
is a vector and $\lambda^l$ acts as a temperature parameter for the softmax
function. We see that increasing weight decay produces smoother probability
distributions, and therefore increases the contribution of correctly
classified images to the final loss.

Adversarial training can similarly be understood as a way of altering the
distributions of penalties over the data during training.
Let $p_y(x)$ be the probability attributed to the true
class $y$. For a perturbation $\delta$, a first order approximation
of $p_y(x + \delta)$ can be written:
$$p_y(x + \delta) \approx p_y(x) + \partial_x\,p_y \cdot \delta$$
and for an $L_2$ constraint, the worse case perturbation is
$\displaystyle \delta = -\epsilon\,\frac{\partial_x\,p_y}
{\|\partial_x\,p_y\|_2}$, yielding:
$$p_y(x + \delta) \approx p_y(x) -\epsilon\,\|\partial_x\,p_y\|_2\;.$$
Training on the perturbed image $x + \delta$ appears equivalent to
training on the image $x$ with a probability attributed to the
correct class that has been decreased by $\epsilon\,\|\partial_x\,p_y\|_2$.
In practice, we reduce the extent to which this process is an approximation
for deep networks by taking multiple steps of gradient descent with a small
step size~\citep{madry2018towards}.

\subsection{Output Regularization}

The idea of regularizing a neural network by altering the way penalties
are distributed over the data during training has been introduced multiple
times before, independently.~\cite{pereyra2017regularizing} refer to this
idea as ``output regularization''.
For instance,~\cite{szegedy2016rethinking} propose to penalize
over-confident predictions with~\emph{label smoothing}, where one-hot
labels are replaced with smooth distributions.~\cite{xie2016disturblabel}
implement a stochastic version of label smoothing with~\emph{disturbLabel},
where labels randomly become incorrect with a small probability.
\cite{pereyra2017regularizing} introduce an alternative to
label smoothing in the form of an entropy-based penalty on high-confidence
predictions. The BANG technique~\citep{rozsa2016towards} discussed in
Section~\ref{sec:related-work} can also be interpreted as an output
regularizer, as it was explicitly designed to increase the contribution
of correctly classified images to the loss during training.
\cite{na2017cascade} and~\cite{kannan2018adversarial}
propose to add a term to the loss to encourage natural images and their
adversarial counterparts to have similar logit representations.
\cite{kannan2018adversarial} also introduce~\emph{logit squeezing},
in which the norm of the logits is penalized.

Based on the considerations of the previous section, we can regard weight
decay and adversarial training as two additional members of this family
of output regularizers. Note that weight decay is particularly closely
related to logit squeezing: the effect of both regularizers is to shrink
the norm of the logits, and incidentally, the Lipschitz constant of the
network.

\section{CIFAR-10 Experiments}
\label{sec:cifar-exp}

\newcommand{\cfwidth}{1.45in}
\newcommand{\cifarwid}{2.6in}
\newcommand{\foolexwidth}{0.40in}
\newcommand{\pgdexpath}{k32-pgd-7-2-8-epoch-250} 
\newcommand{\combexpath}{k32-pgd-7-2-8-epoch-250-l21e-3} 

We now empirically verify previous claims motivated by studying linear models,
in the context of deep learning. We compare a 32-layer
WideResNet~\citep{zagoruyko2016wrn} with width factor 10,
to a simple 4-layer CNN described in Table~\ref{tab:cnn}.
The CNN architecture has 0.4\% as many parameters
as the WideResNet, and we expect its mapping from input to output to
be more linear since it has $8\times$ less depth.
\begin{table}[h]
\centering
\caption{Simple fully-convolutional architecture adapted from the
CleverHans library tutorials~\citep{papernot2017cleverhans}. Model uses
ReLU activations, and does not use batch normalization, pooling, or biases.
We respectively denote $h$, $w$, $c_{in}$, $c_{out}$, $s$
as kernel height, width, number of input and output channels
w.r.t.~each layer, and stride.}
\label{tab:cnn}
\begin{tabular}{|c|c|c|c|c|c|c|} 
\hline 
Layer & $h$ & $w$ & $c_{in}$ & $c_{out}$ & $s$ & params \\
\hline 
\texttt{Conv1} & 8 & 8 & 3 & 32 & 2 & 6.1k\\ \hline 
\texttt{Conv2} & 6 & 6 & 32 & 64 & 2 & 73.7k\\ \hline 
\texttt{Conv3} & 5 & 5 & 64 & 64 & 1 & 102.4k\\ \hline 
\texttt{Fc1} & 1 & 1 & 256 & 10 & 1 & 2.6k\\ \hline 
Total & -- & -- & -- & -- & -- & \textbf{184.8k}\\ \hline 
\end{tabular}
\end{table}

The simple CNN was trained for 250 total epochs with
label smoothing, the Adam optimizer, an initial learning rate of 1e-3,
and batch size of 128. No data augmentation was used other than adversarial
training when indicated, e.g.~by ``PGD (7--2--8)'' for 7 iterations of PGD with
a step size of 2, bounded by $\epsilon_{\max}$ of 8---all out of 255---per
convention for CIFAR-10. This 7--2--8 combination was used
by~\citet{madry2018towards} to secure their models for CIFAR-10,
therefore we use this as default.

We delay adversarial training until after the first 50 epochs to let
the model partially converge so that it can generate ``good''
adversarial examples to train on for the remaining 200 epochs. This is also
sometimes done for efficiency reasons, e.g.~on ImageNet~\citep{kurakin2017adversarial}.
We denote as ``CNN-L2'' and ``CNN'' models trained per the above procedure,
with and without $L_2$ weight decay with regularization constant $\lambda$ =
1e-3.

We opt for 32 filters in~\texttt{conv1}, as opposed to the CleverHans default
of 64, because this was roughly the largest model which could be regularized
s.t.~it was unable to memorize training data examples, while maximizing clean
test accuracy. We verify this by attempting to construct fooling images, and
then evaluate our models on $L_\infty$, and two held-out $L_0$ counting
``norm'' attacks.

\subsection{Fooling Images}

Fooling images~\citep{nguyen2015deep}, also known as ``rubbish
examples''~\citep{goodfellow2015explaining}, are a class of adversarial
examples that begin with an out-of-distribution example, typically sampled from
a Gaussian or uniform distribution, then maximizing the probability of a target
class by accumulating several gradient updates in the input. This procedure has
led to observations that we can easily cause DNNs to predict any chosen
target class with high-confidence, even when the input remains completely
unrecognizable to humans. This method is also motivated by model-based
optimization, where one wants to~\emph{generate} examples that optimize some
criteria, e.g.~objects with a small drag coefficient, given a model that is
only trained to distinguish, or regress over inputs.

When we use 64 filters in~\texttt{conv1} (and $2\times$ this number
in~\texttt{conv2,3}), we get significant filter
sparsity, as both PGD and weight decay encourage this, yet the model can
be coerced into reproducing some training data with fine detail.
In addition to being a poor generalization strategy, this is a security
vulnerability if the training data are confidential e.g.,~users' faces or
banking information, rather than publicly available
images (see~\cite{carlini2018secret} for other ways to extract secrets).
We limit the number of filters in~\texttt{conv1} to 32 to encourage
models to learn a low-dimensional manifold capturing the
data-generating distribution, but not the actual data.
Although the model still has roughly $3\times$ more parameters than CIFAR-10
has examples, it has been shown that we can obtain
parameter-independent capacity control with suitable norm-constraints
for shallow models~\citep{neyshabur2015normbased}.

\begin{figure}[h]
\subfigure[]{
\centering{
\includegraphics[width=3.1in]{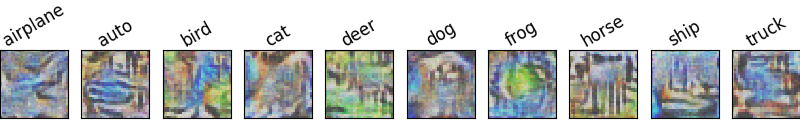} 
\label{fig:fooling-cnn}}} 
\subfigure[]{
\centering{
\includegraphics[width=3.1in]{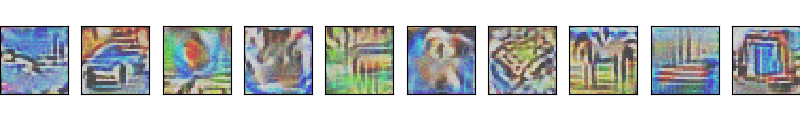} 
\label{fig:fooling-cnn-l2}}} 
\caption{Fooling images for each of the CIFAR-10 classes
using~\subref{fig:fooling-cnn} CNN and~\subref{fig:fooling-cnn-l2}
CNN-L2. Noise is gradually transformed to resemble training data, but
fine detail is omitted. Examples in~\subref{fig:fooling-cnn} are
classified with 32--63\% confidence, and 41--81\%
for~\subref{fig:fooling-cnn-l2}. Best viewed in colour.}
\label{fig:cifar10_cnn_foolex}
\end{figure}

Fooling image samples crafted with the model in Table~\ref{tab:cnn} are shown
in Figure~\ref{fig:cifar10_cnn_foolex}. We sample uniformly from
$0.5 + \mathcal{U}(0, 0.1) \in \mathbb{R}^{32	\times 32 \times 3}$ and do 50
iterations of gradient descent with a step size of 5e-3 on the loss using each
class as the target label for a given trial.

\begin{figure}[h]
\centering{
\includegraphics[width=\cifarwid]{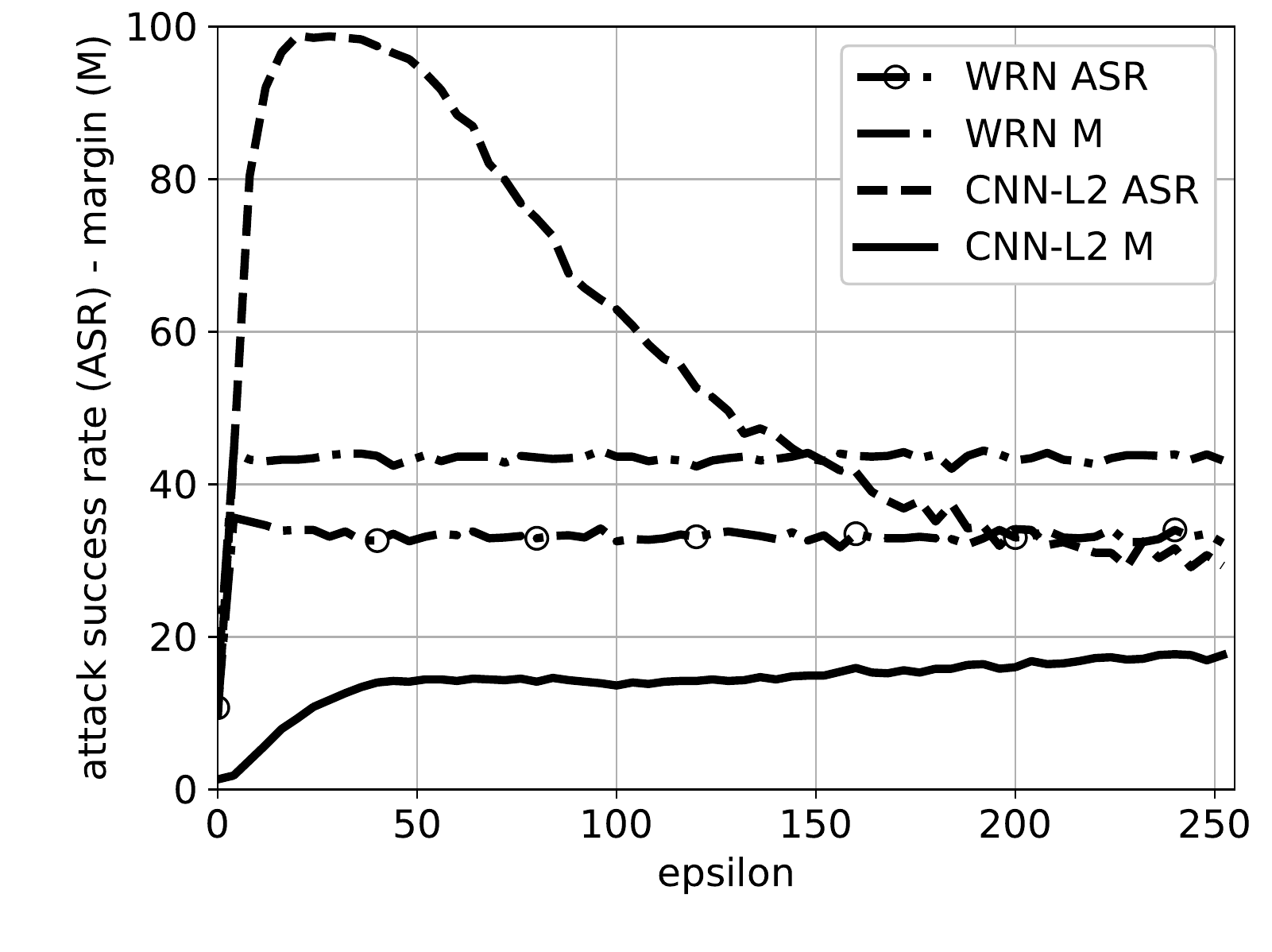}}
\caption{Attack success rate (ASR) and prediction margin (M) for a
targeted single-step fooling images attack on CNN-L2 and the
``Public''~\citeauthor{madry2018towards}~WRN\@. Even though
CNN-L2 is much more linear than WRN, it is better at extrapolating on
challenging out-of-distribution examples and has much smaller margin than WRN\@.}
\label{fig:cifar10_fooling}
\end{figure}

To make the fooling images test more quantitative, we adopt the procedure
suggested by~\citet{goodfellow2015dlss}, of reporting the rate at which
FGSM succeeds in causing the model to hallucinate a target class.
This test is typically not reported in other works as robustness in this
regime remains elusive, and
machine learning models are thought to have a ``thin manifold of accuracy''
due to their linear behaviour which makes them overly confident when
extrapolating~\citep{goodfellow2015dlss, goodfellow2015explaining}.
We consider the attack a success if the argmax logit matches the target
label, or equivalently, if the target softmax probability
exceeds~\nicefrac{1}{10}. This is conservative vs.~choosing the
threshold 0.5 per~\citet{goodfellow2015dlss}.
We sweep $\epsilon$ over the range 0--255, and plot in
Figure~\ref{fig:cifar10_fooling} the attack success rate (ASR), with the
difference between the two largest softmax probabilities the margin (M).
For each $\epsilon$, ASR is an average over 1000 samples from
$\mathcal{U}(0, 1) \in \mathbb{R}^{32 \times 32 \times 3}$, then taking
a step in the direction that maximizes each of the 10 classes.

ASR peaks at around $\epsilon=20$ for CNN-L2, and then decays almost
exponentially with increasing $\epsilon$ as we move away from the manifold.
Meanwhile, the softmax layer is close to uniform and the prediction margin
remains small but slightly increasing. If we use 0.5 as the requisite
prediction threshold, then the ASR is 0\% for $\epsilon=16$ where
it is nearly maximal in Figure~\ref{fig:cifar10_fooling} for the argmax case.
WRN here is the ``Public'' model from the~\citeauthor{madry2018towards}
CIFAR-10
challenge\footnote{\url{https://github.com/MadryLab/cifar10_challenge}}. The
ASR for WRN plateaus after an initial upward spike, as it is very non-linear,
therefore a linear approximation of the loss gets ``stuck''.
Despite this, the WRN is more than twice as confident for all $\epsilon$ than
the CNN\@. It was previously shown that taking multiple steps results
in~\emph{maximum} confidence predictions for the WRN, despite the resulting
examples maintaining a uniform noise appearance~\citep{galloway2018predicting}.

\subsection{$L_\infty$ Attacks}

\begin{figure}[h]
\centering{
\includegraphics[width=\cifarwid]{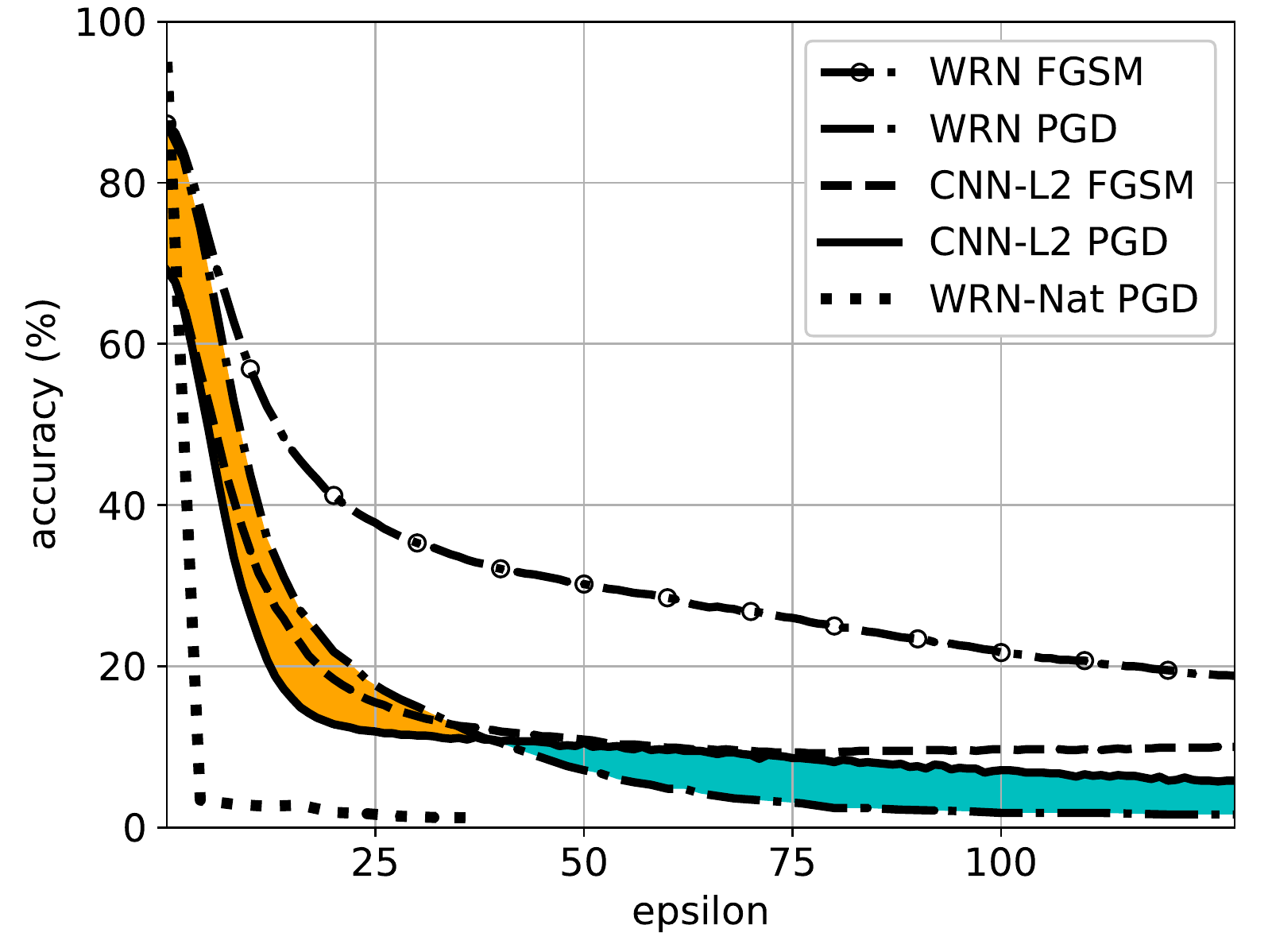}}
\caption{White-box $L_\infty$ optimized gradient-based attacks.
The cyan-fill area is where CNN-L2 achieves higher accuracy than WRN
(the~\citeauthor{madry2018towards} defense), and the orange area is
vice versa. The ``PGD'' attack is the (20--2--$\epsilon$) variant
using the cross entropy loss and random start. FGSM is only shown for
context to compare the marginal improvement of the PGD attack.}
\label{fig:cifar10_pgd_attk}
\end{figure}

In Figure~\ref{fig:cifar10_pgd_attk} we compare our CNNs to the ``Public'' and
``Secret'' pre-trained models that currently hold the benchmark for robustness
to $L_\infty$ attacks in the white-box and black-box setting respectively.
These two models arise from independent runs of the same procedure, therefore
their curves were almost identical over the full range of $\epsilon$ for
gradient sign attacks. We average the two series and denote ``WRN'' to reduce
clutter. A baseline ``WRN-Nat'' trained by~\citeauthor{madry2018towards}
without PGD is included for context. As in their Figure 2, we sweep over
$\epsilon$ for white-box attacks, but continue on to $\epsilon_{\max}=128$
until all predictions were reduced to chance.

We do not equate~\emph{accuracy} to \emph{robustness} for white-box attacks,
since the original adversarial examples problem also considers the saliency of
perturbations. Figure~\ref{fig:cifar10_cnn_foolex} depicts how gradient-based
perturbations with our models tend to be highly recognizable and structured for
targeted attacks. We expect the~\citeauthor{madry2018towards}~models to
generally have higher accuracy for the $L_\infty$ attacks characterized in
Figure~\ref{fig:cifar10_pgd_attk}, since they allocate significant capacity for
this express purpose. However, it is interesting to note that our small models
eventually~\emph{do} overtake the them for $\epsilon > 40$. It is also
encouraging that the slopes remain similar between ``CNN-L2 PGD'' and ``Madry
et al. PGD'', with the offset staying roughly constant in the small $\epsilon$
regime. Neither model fails as catastrophically as the undefended WRN-Nat.

We provide curves for FGSM not as a meaningful attack, but to show that there was
little marginal benefit of performing multiple attack iterations for CNN-L2,
compared to WRN\@. This shows that CNN-L2 is more linear~\emph{locally} near
natural data manifold---as well as~\emph{globally} with fooling images---than
the WRN\@.

\subsection{$L_0$ Attacks}

The $L_0$ counting ``norm'' was previously proposed as a natural metric
for measuring a model's ability to resist perturbations~\citep{papernot2016distillation}.
We now evaluate on two held out $L_0$ attacks: random pixel swapping, and the
Jacobian-based Saliency Map Attack (JSMA)~\citep{papernot2016limitations} as
additional benchmarks. For both of these attacks, each RGB colour channel was perturbed
separately.

\subsubsection{Pixel Swapping}
\label{{sec:pixel-swapping}}

Randomly swapping pixels can be interpreted as a weak black-box attack
that controls for the perturbations' saliency. This type of perturbation may be less
detectable than JSMA by some statistical methods, as it does not affect the data
distribution globally, however most detection methods do not work well~\citep{carlini2017bypassing,athalye2018obfuscated,smith2018understanding}.

\begin{figure}[h]
\centering{
\includegraphics[width=\cifarwid]{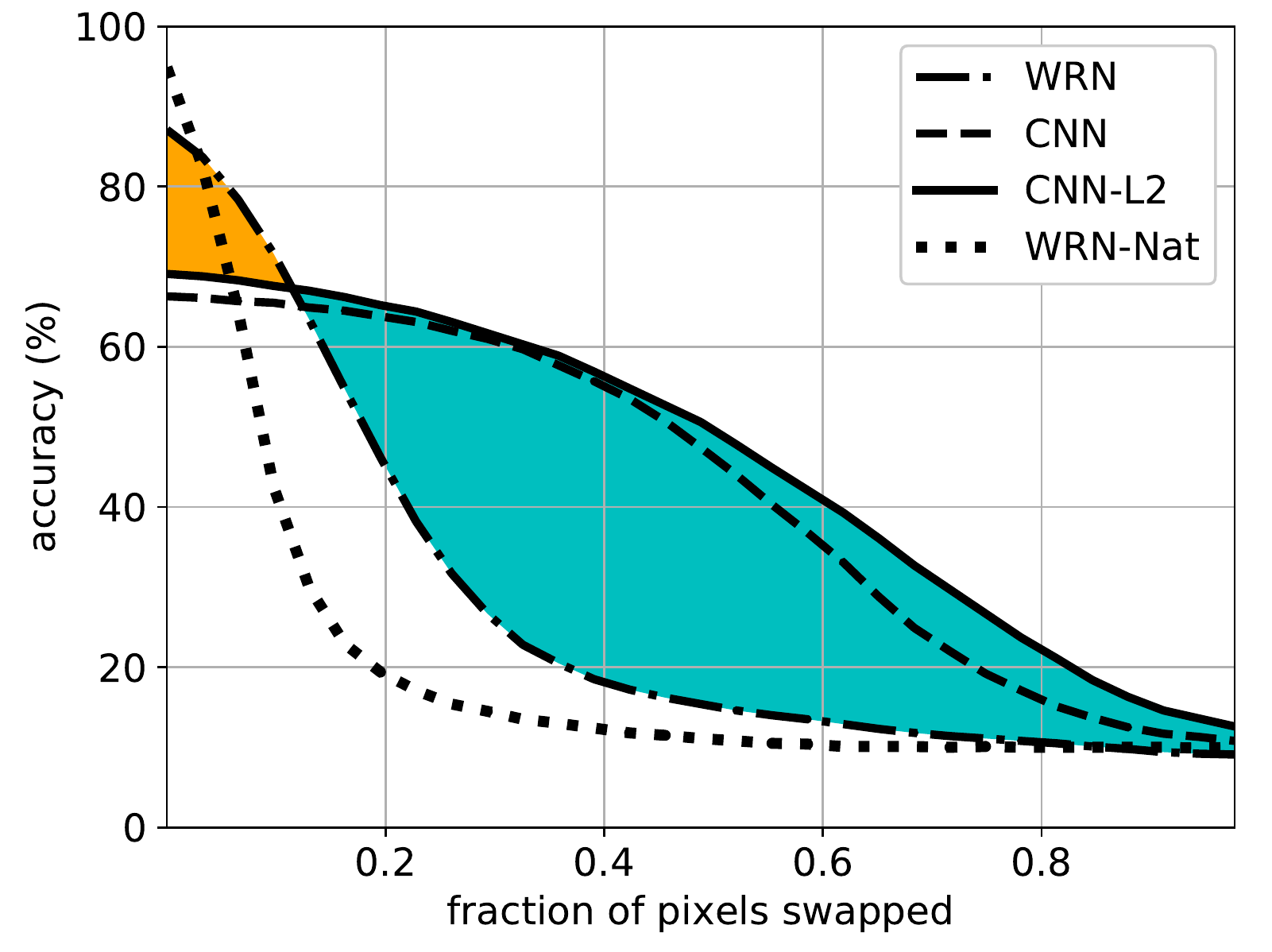}}
\caption{An $L_0$ counting ``norm'' black-box attack in which we randomly
swap pixels, and R,G,B are considered separately.
The cyan-fill area is where CNN-L2 is more robust than
the~\citeauthor{madry2018towards}~defense, and the orange area is vice
versa.}
\label{fig:cifar10_rand_swap_attk}
\end{figure}

\begin{figure}[h]
\centering{
\includegraphics[width=0.45\textwidth]{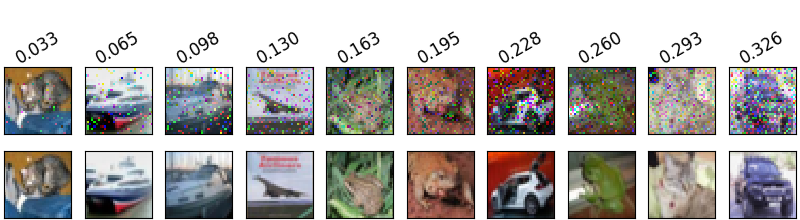}}
\caption{CIFAR-10 samples with ratio of pixels randomly swapped
annotated (\emph{top}) and original images
(\emph{bottom}). At 26\% of pixels swapped, CNN-L2 achieves 63.1\%
accuracy, WRN--31.7\%, and WRN-Nat--15.4\%.}
\label{fig:cifar10_rand_ex}
\end{figure}

In Figure~\ref{fig:cifar10_rand_swap_attk}, we plot accuracy for four different
models, with each series averaged over three runs. Despite the attack being
random, variance was very low even when not setting the random seed. CNN-L2
overtakes the WRN beyond $\approx 11\%$ of features swapped,
and WRN-Nat at $\approx 5\%$. For context, Figure~\ref{fig:cifar10_rand_ex}
shows sample images with decreasing signal-to-noise (we did not normalize to
an actual SNR). The noise is fairly easily detected by human vision when presented
next to the original, but the overall image remains interpretable.

\subsubsection{Jacobian-based Saliency Map Attack}
\label{sec:jsma}

Gradient-based attacks tuned for the $L_0$ metric have also been proposed,
such as JSMA which uses a greedy iterative procedure that constructs a saliency
map, and sets pairs of pixels to the full-scale value until the attack either
succeeds, or some maximum number of pixels are modified. This maximum is usually
expressed as a ratio of total pixels, as in Section~\ref{{sec:pixel-swapping}}.

In Table~\ref{tab:jsma} we report JSMA targeted success rates when the adversary
may perturb between 1--5\% of features, denoted by $\gamma$. CNN-L2 limits
the adversary to lower attack success rates for all $\gamma$ in this range,
even when considering its lesser clean test accuracy relative to WRN, as we
do not skip examples that are incorrectly classified originally.

\begin{table}[]
\centering
\caption{JSMA success rate (ASR), avg.~rate of perturbed features overall
(Pert.~Rte), and when attack was successful (S Pert.~Rte.) for a number of
source images N. Maximum ratio of features that may be perturbed denoted by
$\gamma$ (as a percentage). For each source image, a targeted attack is
performed using all non-true labels as the target.}
\label{tab:jsma}
\begin{tabular}{|c|c|c|c|c|} 
\hline 
\rowcolor{Gray}
\mc{5}{WRN} \\ \hline 
N & $\gamma$ & ASR & Pert. Rte. & S Pert. Rte. \\ \hline 
\multirow{5}{*}{10} & 1\% & 33\% & 0.8\% & 0.19\% \\ \cline{2-5}
 & 2\% & 58\% & 1.3\% & 0.51\% \\ \cline{2-5}
 & 3\% & 81\% & 1.5\% & 0.99\% \\ \cline{2-5}
 & 4\% & 86\% & 1.7\% & 1.14\% \\ \cline{2-5}
 & 5\% & 89\% & 1.8\% & 1.28\% \\ \hline 
100 & 1\% & 28.4\% & 0.8\% & 0.14\% \\ \hline 
\hline 
\rowcolor{Gray}
\mc{5}{CNN-L2} \\ \hline 
\multirow{5}{*}{10} & 1\% & \textbf{20\%} & 0.8\% & 0.11\% \\ \cline{2-5}
 & 2\% & \textbf{52\%} & 1.4\% & 0.51\% \\ \cline{2-5}
 & 3\% & \textbf{69\%} & 1.7\% & 0.88\% \\ \cline{2-5}
 & 4\% & \textbf{81\%} & 1.9\% & 1.25\% \\ \cline{2-5}
 & 5\% & \textbf{88\%} & 2.1\% & 1.52\% \\ \hline 
100 & 1\% & \textbf{26.9\%} & 0.8\% & 0.13\% \\ \hline 
\end{tabular}
\end{table}

\subsection{Discussion}

It can be observed in Figure~\ref{fig:cifar10_rand_swap_attk} that $L_2$ weight
decay had a beneficial effect in defending against the $L_0$ attack for all
$\epsilon$, with the benefits being most pronounced when the ratio of pixels
swapped was less than 0.2, or between 0.4 and 0.9.

We omitted ``CNN'' in Figure~\ref{fig:cifar10_pgd_attk} to keep the figure
clear, and offer a comparison between with ``CNN-L2'' here instead. For the PGD
(20--2--$\epsilon$) attack, the curves have similar slope for $\epsilon < 16$,
with that of CNN being translated downward relative to CNN-L2. The
difference was usually between 3--6\% for this region, with the greatest difference
of 6.8\% at $\epsilon=5$. Beyond $\epsilon=16$, CNN was better, but
accuracy was already quite low at this point around $\approx 15\%$. We found
that the more a model underfit w.r.t.~clean test accuracy (e.g.~by optimizing with
SGD instead of Adam, or binarizing hidden representations) the more gradual the
roll-off when attacking with PGD, and~\emph{earlier} the cross-over point
vs.~the WRN\@.

\begin{figure}[h]
\centering{\subfigure[]{\includegraphics[width=\cfwidth]{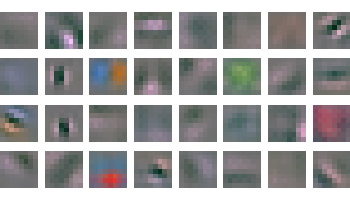}
\label{fig:conv1_pgd}}} 
\centering{\subfigure[]{\includegraphics[width=\cfwidth]{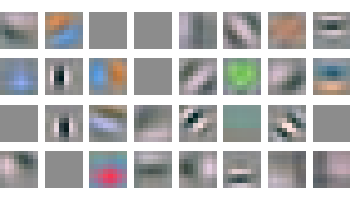}
\label{fig:conv1_pgd_wd}}} 
\caption{Visualizing filters from layer~\texttt{conv1} for the
model outlined in Table~\ref{tab:cnn}. Filters arising from
training~\emph{without} weight decay in~\subref{fig:conv1_pgd} have
a highly textured effect, while unimportant filters go to zero
in~\subref{fig:conv1_pgd_wd} with overall $L_2$ weight decay.
Both models were trained with the same random seed, hence why
similar filters occur at the same coordinates in the grid.
Best viewed in colour.}
\label{fig:cifar10_filters}
\end{figure}

Visualizing the filters of CNN-L2 in Figure~\ref{fig:cifar10_filters}
further speaks to the generality of this model.
Given the fundamentally limited diversity and finite number of examples in
CIFAR-10, it is desirable to learn a distributed representation comprising
simple image primitives such as edge and colour detectors. These types of
filters emerge naturally as a result of training a small architecture for
a long time (e.g.~250 epochs), however models trained with weight decay
(Figure~\ref{fig:conv1_pgd_wd}) have smoother filters, with some unnecessary
ones becoming exactly zero.

The random swap experiment is sufficient to demonstrate that our models are
significantly more robust to at least one attack defined w.r.t the $L_0$
metric. We believe it is unlikely that the~\citeauthor{madry2018towards}
``Public'' or ``Secret'' models will perform in a more compelling manner than
``CNN-L2'' over a similar range for~\emph{any} $L_0$ attack, but we invite
others to attempt to show this by attacking our pre-trained model. Note that no
specific action was taken to confer performance for $L_0$ attacks during training,
other than attempting to tune a model to generalize well.

Models trained on PGD adversarial examples artificially increase the margin
between class centroids by translating them with respect to the loss
function, as PGD is an equivalent $L_\infty$ analogue
to FGSM for deeper models. This reduces data separability
by $\approx \epsilon \|w\|_1$ when the signed gradient is used,
which encourages the model to fit some dimensions of low-variance
in the training data to fully separate the classes.
This results in the highly textured effect seen in Figure~\ref{fig:conv1_pgd},
supporting previous observations that CNNs have a tendency to latch on
to surface statistics to fulfill the ERM optimization objective,
resulting in a narrow generalization~\citep{jo2017measuring}.

The tendency of adversarial training to encourage fitting
a specific metric, to a given magnitude, is even more pronounced
with the harsh cliff in~\citeauthor{madry2018towards}
(their Figures 2 a) and b)) for MNIST\@. We omit an MNIST 
comparison for brevity, but refer readers to examples of targeted
hand-draw attacks in a mobile
application\footnote{\url{https://github.com/AngusG/tflite-android-black-box-attacks}}.

\section{Conclusion}

As we may not anticipate~\emph{how} our systems are likely to
be compromised, a reasonable choice is to generalize
to the best of our ability given finite data.
Adversarial training has a limited ``sweet-spot'' in
which it is effective at increasing robustness to~\emph{some}
attacks, whereas weight decay reduces all generalization errors
and encourages underfitting in the worst-case.
Adversarial training is effective when the perturbation size is small,
but remains challenging to apply without exposing other vulnerabilities.
By significantly reducing the hypothesis space, our models strike compelling
trade-offs when subject to unseen attacks.
This provides a more complete baseline for future robustness comparisons
which may involve more sophisticated learning objectives, and
regularization techniques that do not degenerate with depth.
Incorporating prior expert knowledge and interpretability measures
seem promising for limiting the impact of adversarial examples, given sparsely
sampled datasets in high-dimensions. We suggest
future works sweep over  the full perturbation range for several unseen attacks
before drawing robustness conclusions.

\subsubsection*{Acknowledgments}

The authors wish to acknowledge the financial support of NSERC, CFI, CIFAR and
EPSRC\@. The authors also acknowledge hardware support from NVIDIA and Compute
Canada. We thank Colin Brennan, Brittany Reiche, and Lewis Griffin for helpful
edits and suggestions that improved the clarity of our manuscript.


\bibliography{master}
\bibliographystyle{abbrvnat}
\setcitestyle{authoryear,open={(},close={)}} 

\end{document}